\documentclass[conference]{IEEEtran}
\IEEEoverridecommandlockouts
\usepackage{cite}
\usepackage{amsmath,amssymb,amsfonts}
\usepackage{algorithmic}
\usepackage{graphicx}
\usepackage{textcomp}
\usepackage{xcolor}
\usepackage{subcaption}

\usepackage{paralist}

\usepackage{amsthm}

\usepackage{fancyhdr}

\def\BibTeX{{\rm B\kern-.05em{\sc i\kern-.025em b}\kern-.08em
    T\kern-.1667em\lower.7ex\hbox{E}\kern-.125emX}}

\newcommand{\mytitle}{Presented at the Workshop on Uncertainty in Open World Robotics of the 2026 IEEE International Conference on Robotics \& Automation.}
  \newcommand{\figmargin}{-5mm}
     \newcommand{\extrasmall}{\fontsize{8}{8}\selectfont}

\title{All Models are Wrong, Knowing Where is Useful:\\ On Model Uncertainty in Reinforcement Learning
}
 \author{Bernd Frauenknecht, Devdutt Subhasish, Artur Eisele, Friedrich Solowjow, and Sebastian Trimpe
 \thanks{This work is funded in part by the German Federal Ministry of Research, Technology and Space (BMFTR) under the Robotics Institute Germany (RIG), which the authors gratefully acknowledge. Friedrich Solowjow is supported by the KI-Starter grant by the state of NRW.  Further, the authors gratefully acknowledge the computing time provided to them at the NHR Center NHR4CES at RWTH Aachen University (project number p0022301).}
\thanks{Institute for Data Science in Mechanical Engineering, RWTH Aachen University, 52062 Aachen, {\tt\extrasmall firstname.name@dsme.rwth-aachen.de}}
 }

\begin{document}
\maketitle
\thispagestyle{fancy}
\pagestyle{empty}

\begin{abstract}
Model-based reinforcement learning (MBRL) infers information about the environment from a learned dynamics model and bears the potential to address open problems such as data efficient and safe learning in robotics.
However, inaccuracies of the learned dynamics model are typically exploited by the agent, substantially hampering the capabilities of MBRL methods. We present a framework for dealing with inaccuracies of probabilistic models through targeted handling of uncertainty that effectively mitigates model exploitation. We present recent successes in learning directly on hardware and safe exploration, and discuss future directions for uncertainty-aware MBRL.
\end{abstract}


\section{Introduction}

Deep reinforcement learning (RL) has achieved remarkable successes in robot learning problems such as drone racing \cite{kaufmann2023champion} and legged locomotion \cite{rudin2022learning}. However, data inefficiency of state-of-the-art RL approaches and their unsafe exploration behavior remain key challenges for real-world applications \cite{Tang2025May}. Consequently, the common approach to RL in robotics is pretraining in a physics simulator with domain randomization and subsequent sim-to-real transfer \cite{kaufmann2023champion, rudin2022learning}. However, implementing a fast and accurate simulator with a suitable randomization is laborious or even impossible, e.g., for contact-rich problems.

Model-based reinforcement learning (MBRL) is a common approach to address data efficiency \cite{frauenknecht_data-efficient_2023,Chua2018} by learning a dynamics model to infer information without interacting with the real world. A model can further be used to plan ahead and avoid dangerous actions. While this conceptually enables efficient and safe learning, MBRL suffers from model exploitation in practice.
 That is, the RL agent exploits inaccuracies in the dynamics model, leading to poor performance on the real system. In this position paper, we

\begin{figure}[tb]
\centerline{\includegraphics[width=\columnwidth]{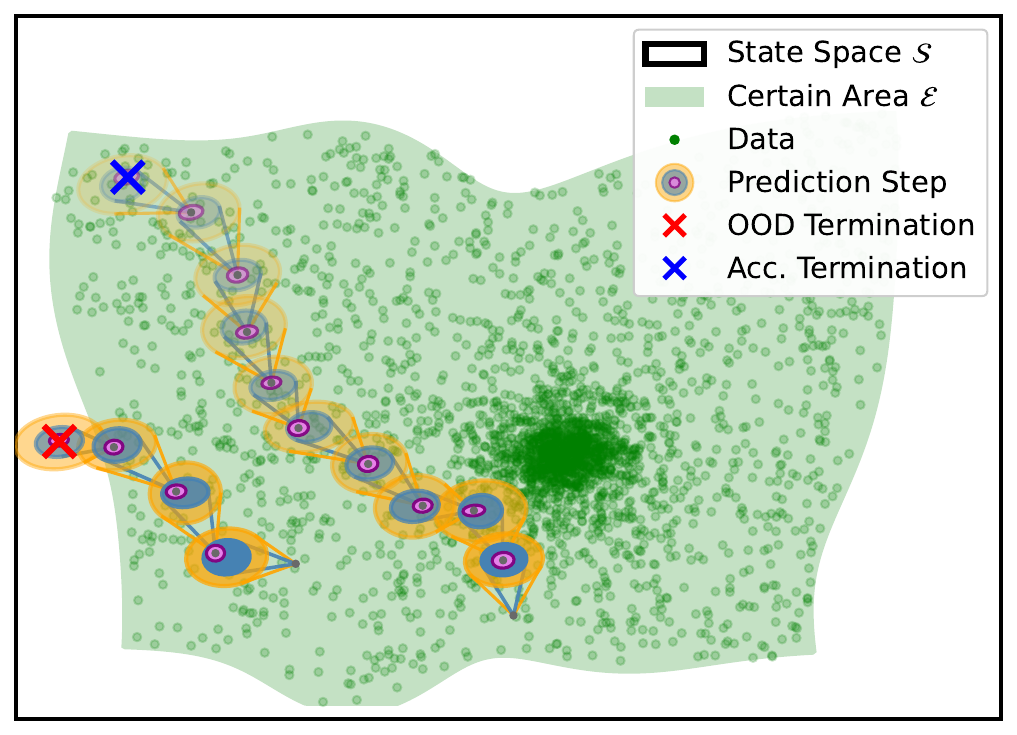}}
\caption{Uncertainty-Aware Model Usage. \textit{Environment data induces a certain area $\mathcal{E}$ of the model where it is accurate. Infoprop model rollouts terminate when they leave $\mathcal{E}$ and go out-of-distribution (red) or when small errors (increasing transparency) have accumulated over long horizons (blue). }}
\label{fig:fig1}
\vspace{\figmargin}
\end{figure}

\begin{compactitem}
\item[(i)] develop an overarching perspective on handling model uncertainty based on methods originally developed in \cite{frauenknecht2024trust} and \cite{frauenknechtrollouts2025} to mitigate model exploitation (see Fig. \ref{fig:fig1});
\item[(ii)] present recent breakthroughs in learning on hardware \cite{subhasish2026} and safe exploration \cite{eisele2026, frauenknecht2026} leveraging this concept; and
\item[(iii)] highlight directions for future work that could be addressed with this framework.
\end{compactitem}
In a nutshell, we illustrate that careful handling of model uncertainty can substantially enhance the capabilities of MBRL approaches, making them promising candidates for addressing open problems in robot learning.

\section{Background}
We briefly present the fundamentals of model usage in RL.
\paragraph{Reinforcement Learning}
Sequential decision-making problems can be addressed using RL, where an agent interacts with an environment, typically modeled as a Markov decision process (MDP) described by the tuple $\mathcal{M} = \{ \mathcal{S}, \mathcal{A}, r, p, \gamma, \rho_0  \}$. Transitions between states $s_t \in \mathcal{S}$ via actions $a_t \in \mathcal{A}$ are described by the dynamics $s_{t+1} \sim p(\cdot|s_t, a_t)$. During transitions, rewards $r_t \in \mathbb{R}$ are emitted based on a reward function $r_{t+1} = r(s_t, a_t)$. Starting from an initial state $s_0 \sim \rho_0$, the agent aims to find a policy $a_{t} \sim \pi(\cdot | s_t)$ that maximizes the expected discounted sum of rewards, referred to as return $\eta$. Thus, the optimal policy is given by
\begin{equation}
    \pi^* = \arg\max_\pi \eta[\pi] = \arg\max_\pi \mathbb{E}_{\pi, p} \left[\sum_{t=0}^\infty \gamma^t r_{t+1}\right].
    \label{eq:rl_objective}
\end{equation}
\paragraph{Model-Based Reinforcement Learning}
While the environment dynamics $p$ are typically unknown, transition data collected during learning can be used to train a data-driven model $\hat{p}$ of the environment. This model can subsequently be used to infer information without costly or dangerous environment interaction. There are four main categories of MBRL: \emph{(i)} planning methods \cite{Williams2017May, Chua2018,  Hafner2019May} do not learn an explicit policy but iteratively plan their actions in the model, \emph{(ii)} analytic gradient methods \cite{Deisenroth2011Jun, Hafner2020Apr} backpropagate through model-based rollouts to optimize the policy, \emph{(iii)} value expansion methods \cite{Feinberg2018Feb, Buckman2018Dec} improve value estimates through model-based rollouts, and \emph{(iv)} Dyna-style approaches \cite{Sutton1991Jul, janner_when_2019-1} use model rollouts to generate training data for a model-free RL agent. In the following, we focus on the latter as they yield state-of-the-art performance for problems with proprioceptive state representations.

\paragraph{Dynamics Models}
The model class approximating $p$ is a fundamental design decision in MBRL. While early works focus on Gaussian processes \cite{Deisenroth2011Jun} and local linear models \cite{Gu2016Jun}, complex dynamical systems require models with higher representational capacity. 
Since deterministic neural network (NN) models \cite{Williams2017May, Nagabandi2018} tend to overfit sparse initial datasets, Bayesian NNs \cite{Depeweg2016May, gal2016improving} are generally preferred due to their ability to capture predictive uncertainty.
A particularly successful architecture is the probabilistic ensemble (PE) model \cite{Lakshminarayanan2017Dec}, comprising $E$ probabilistic neural networks (PNN). Each ensemble member $e$ is trained on a bootstrapped version of the data and predicts the next state as a Gaussian distribution
\begin{equation}
    \hat{p}^e (\cdot|s_t, a_t) = \mathcal{N}(\cdot|\mu^e(s_t, a_t), \Sigma^e(s_t, a_t)).
\end{equation}
Here, aleatoric uncertainty, due to process noise, is captured in the individual covariance predictions $\Sigma^e$. Epistemic uncertainty, due to lack of data, is reflected in disagreement of the PNN predictions. While aleatoric uncertainty is a property of $p$, epistemic uncertainty vanishes in the limit of infinite data.

PE models are typically propagated using the Trajectory Sampling (TS) rollouts \cite{Chua2018}, where the predicting ensemble member is uniformly sampled for each prediction step.\footnote{To be precise, this is referred to as TS1 in \cite{Chua2018}.} This yields a predictive distribution of the PE to be the Gaussian mixture distribution over its ensemble members
\begin{equation}
    \hat{p}(\cdot | s_t, a_t) = \frac{1}{E} \sum_{e=1}^E \mathcal{N}(\cdot|\mu^e(s_t, a_t), \Sigma^e(s_t, a_t)).
    \label{eq:ts_dist}
\end{equation}
\section{Mitigating Model Exploitation Through Uncertainty Awareness}
\label{sec:uncertainty_quantification}
Uncertainty-awareness is a topic of increasing interest in MBRL, where predictions with high epistemic uncertainty are down-weighted, penalized, or discarded \cite{Buckman2018Dec, Yu2020, Li2025Apr} as this indicates model inaccuracy.
The problem of model exploitation is typically formalized as an overestimation of the expected return under the model dynamics upper bounded by $C$
\begin{equation}
    \eta(\pi) \geq \hat{\eta}(\pi) - C
    \label{eq:model_exp}
\end{equation}
where $\hat{\eta}(\pi) = \mathbb{E}_{\pi, \hat{p}} [\sum_{t=0}^\infty \gamma^t r_{t+1}]$. That is, the agent shows behavior that yields high return under the inaccurate model dynamics but returns in the environment are up to $C$ lower. Consequently, keeping $C$ small mitigates model exploitation.

\subsection{Where To Trust The Model?}
\label{subsec:macura}
Unsurprisingly, $C$ depends on the dynamics misalignment between the model and the environment \cite{luo2019, janner_when_2019-1}, such that return estimates deteriorate if the dynamics model cannot represent environment behavior.
A spatial perspective on model usage is provided by the bound presented by \cite{frauenknecht2024trust}, where
\begin{equation}
    C \propto \sum_{t=0}^T \gamma^t \sum_{k=0}^t\sup_{(s_k,a_k) \in \mathcal{E}} D_{\mathrm{TV}} (p(\cdot |s_k, a_k) \| \hat{p}(\cdot|s_k, a_k)))
    \label{eq:eror_bound}
\end{equation}
depends on the worst case dynamics misalignment within a subset of the state action manifold $\mathcal{E} \subseteq \mathcal{S} \times \mathcal{A}$ in which the model is leveraged, and the considered time horizon $T$. The dynamics misalignment is measured in the total variation distance $D_{\mathrm{TV}}$ between the respective dynamics kernels $p$ and $\hat{p}$. A common way to limit $C$, is keeping horizons $T$ short. This, however, is often undesirable as it may lead to myopic decisions. In contrast, \eqref{eq:eror_bound} indicates that using the model in a set $\mathcal{E}$ with sufficiently low dynamics misalignment is an alternative approach to mitigate model exploitation.

\begin{figure*}[tb]
     \centering
     \begin{minipage}[b]{0.23\textwidth}
         \centering
         \includegraphics[width=\textwidth]{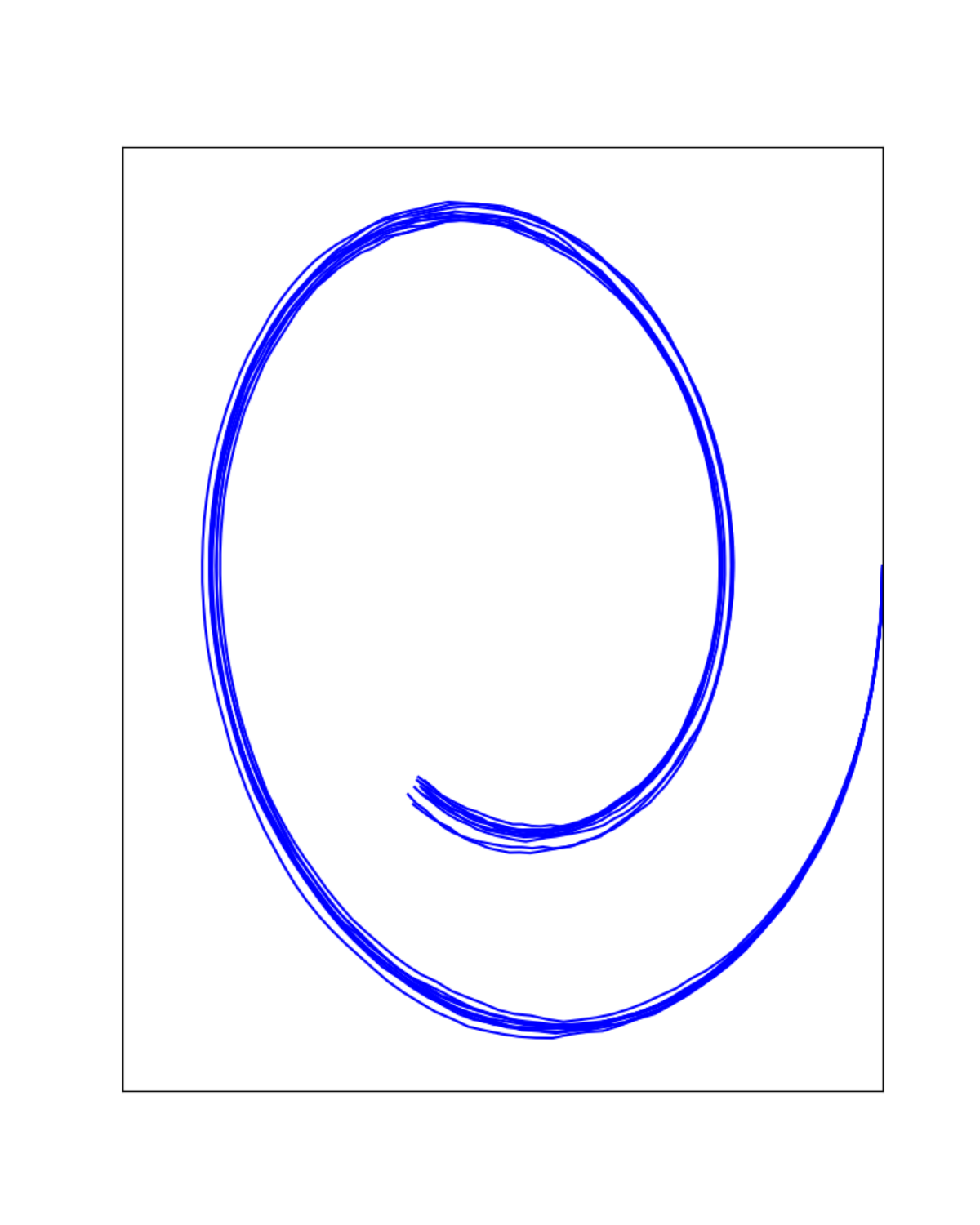}
         \par\smallskip
         (a) $\text{Data}$
     \end{minipage}
     \hfill
     \begin{minipage}[b]{0.23\textwidth}
         \centering
         \includegraphics[width=\textwidth]{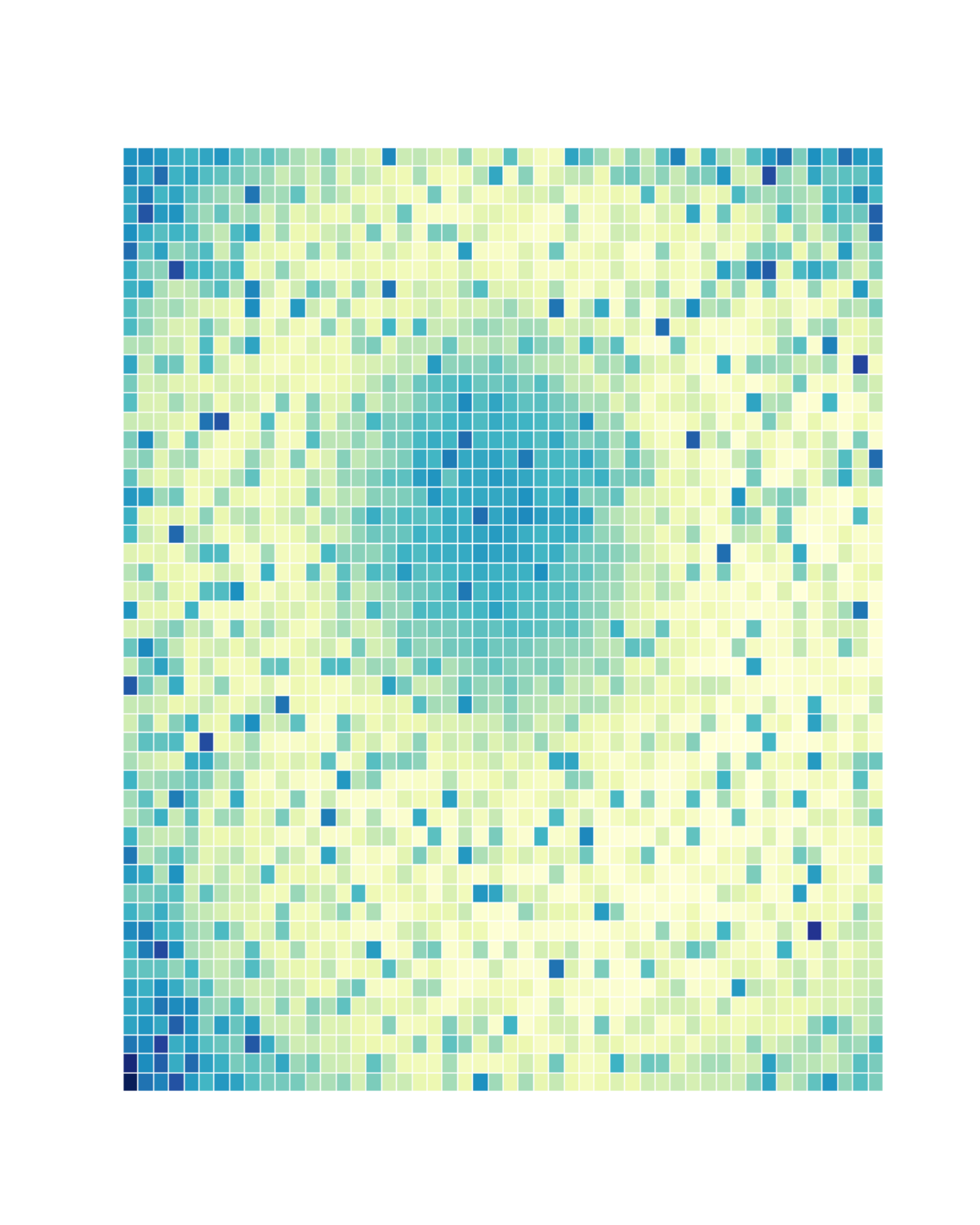}
         \par\smallskip
         (b) $D_{\mathrm{TV}}(p \| \hat{p})$
     \end{minipage}
     \hfill
     \begin{minipage}[b]{0.23\textwidth}
         \centering
         \includegraphics[width=\textwidth]{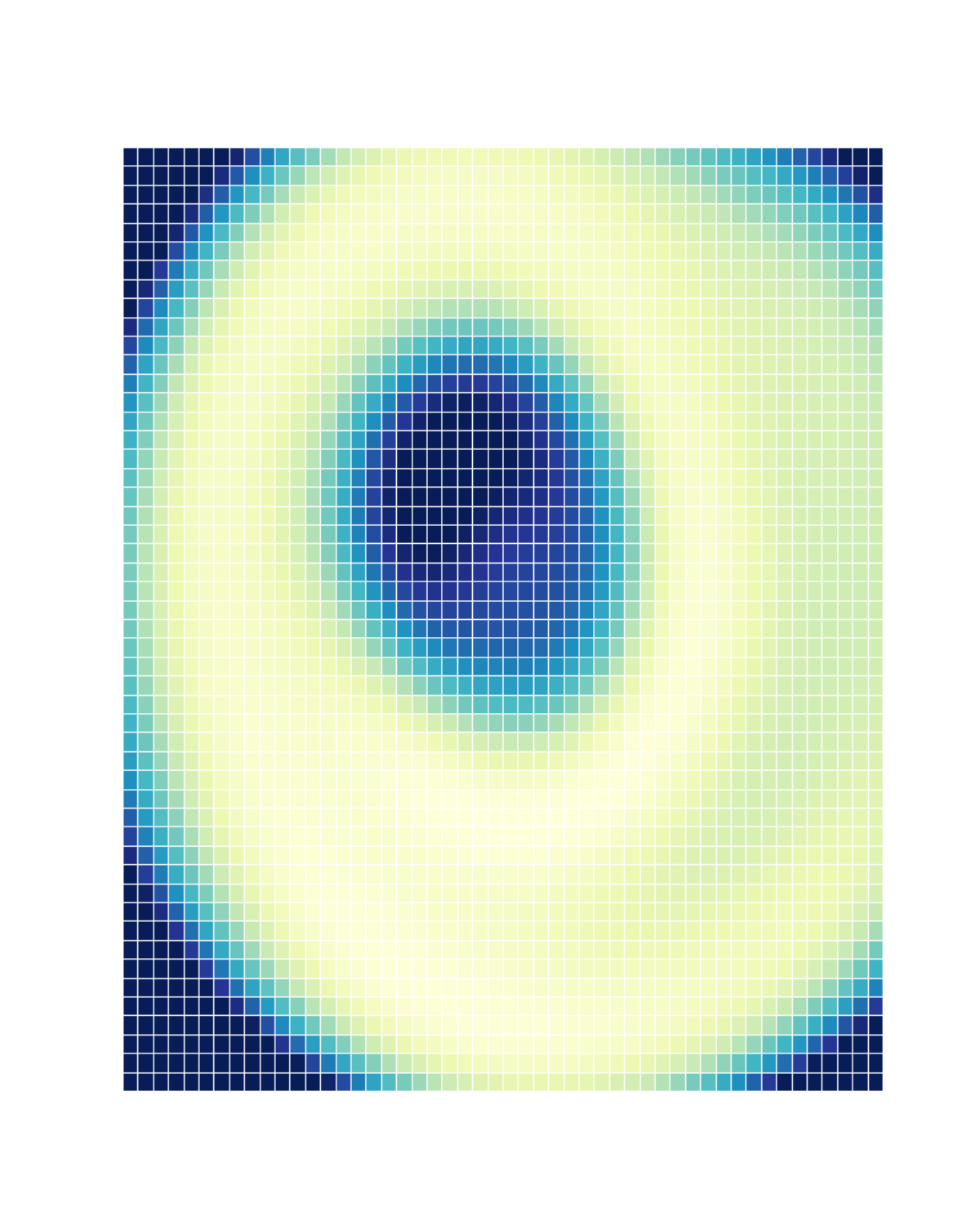}
         \par\smallskip
         (c) $u_{\mathrm{epi}}$
     \end{minipage}
     \hfill
     \begin{minipage}[b]{0.23\textwidth}
         \centering
         \includegraphics[width=\textwidth]{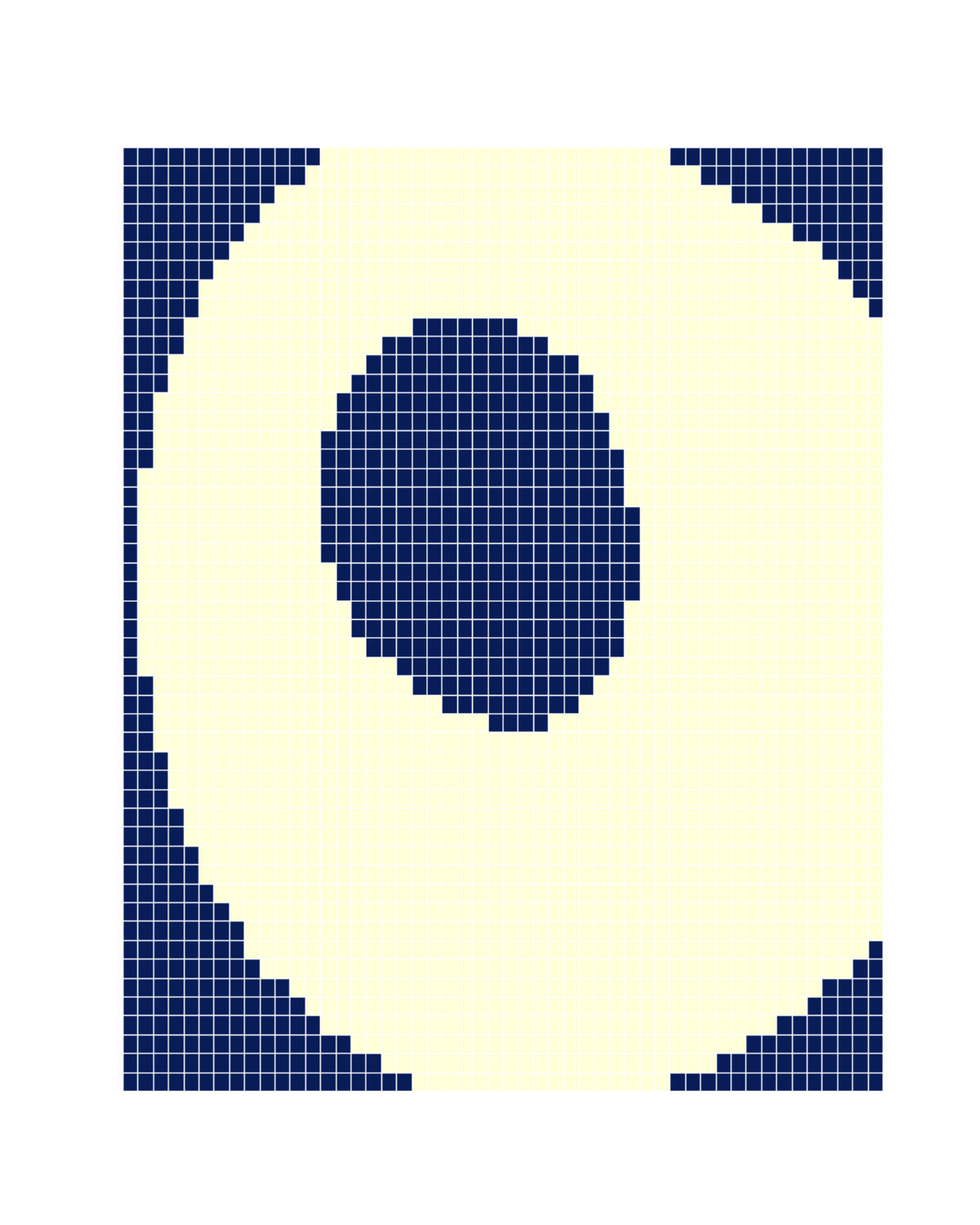}
         \par\smallskip
         (d) $\mathcal{E}$ with $u_{\mathrm{epi}} < \kappa_1$
     \end{minipage}

     \caption{ (from \cite{frauenknecht2024trust}) Constructing $\mathcal{E}$. \textit{(a) Training Data. (b) Dynamics Misalignment. (c) Epistemic Uncertainty. (d) Sufficiently certain set $\mathcal{E}$. Uncertainty is a valid proxy for misalignment. Thus, restricting model usage to $\mathcal{E}$ limits model exploitation.}}
     \label{fig:uncertainty_measures}
     \vspace{\figmargin}
\end{figure*}

Since $p$ is unknown in practice, dynamics misalignment cannot be measured directly but needs to be estimated via a proxy.
Assuming sufficient model capacity, epistemic uncertainty is an expressive indicator of dynamics misalignment as illustrated in Fig. \ref{fig:uncertainty_measures} with a detailed description in \cite{frauenknecht2024trust}. Training a PE model on the data distribution in Fig. \ref{fig:uncertainty_measures}a yields dynamics misalignment depicted in Fig. \ref{fig:uncertainty_measures}b, where blue indicates high and yellow indicates low misalignment. As expected the model represents the environment behavior well close to the training data but deteriorates in out-of-distribution settings. The epistemic uncertainty depicted in Fig. \ref{fig:uncertainty_measures}c shows a similar behavior where yellow corresponds to low and blue to high uncertainty. Consequently, 
restricting model usage to an area with epistemic uncertainty below a threshold $\kappa_1$, yields low dynamics misalignment.
 Thus, we define the certain area
 \begin{equation}
     \mathcal{E} := \{ (s, a) \in \mathcal{S} \times \mathcal{A} | u_{\mathrm{epi}}(s,a) \leq \kappa_1 \} 
     \label{eq:cert_set}
 \end{equation}
 as the subset of $\mathcal{S} \times \mathcal{A}$, where epistemic uncertainty is sufficiently low and, truncate model rollouts that leave $\mathcal{E}$, as indicated in the out-of-distribution termination in Fig \ref{fig:fig1}.
 
 There are different valid options to estimate epistemic uncertainty. For example \cite{Lakshminarayanan2017Dec} propose the average Kullback Leibler divergence between the individual PNN distributions and the PE Gaussian mixture distribution, while \cite{frauenknecht2024trust} choose a pair-wise comparison with a closed form solution.

\subsection{How To Propagate The Model Where It Can Be Trusted?}
\label{subsec:infoprop}
While selecting $\mathcal{E}$ mitigates model exploitation, it does not yet enable planning over long horizons $T$, as small errors can accumulate to prohibitively large deviations. A further step is improving the predictive distribution of the model to reduce dynamics misalignment and estimating error accumulation.

\begin{figure}[b]
\centerline{\includegraphics[width=\columnwidth]{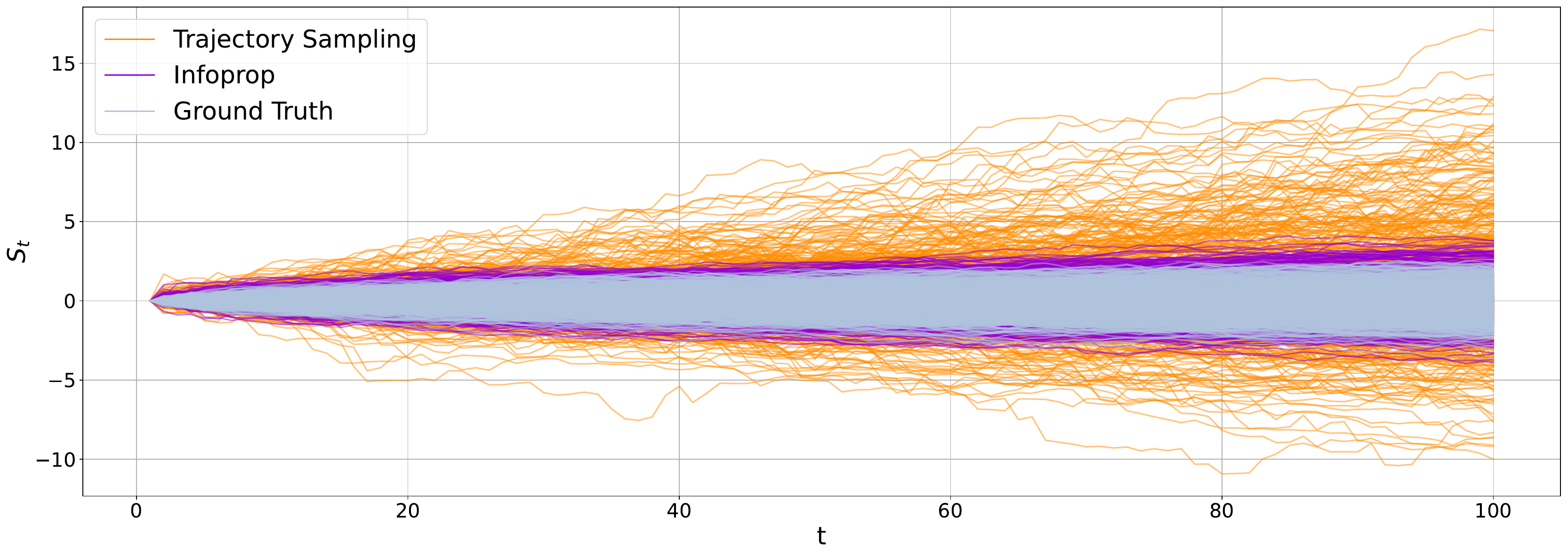}}
\caption{(from \cite{frauenknechtrollouts2025}) Improving the Predictive Distribution. \textit{Infoprop rollouts (purple) closely capture the environment behavior (blue), while TS rollouts (orange) systematically overestimate stochasticity as model errors accumulate.}} 
\label{fig:ip_rw}
\end{figure}

The model's predictive distribution is typically corrupted by epistemic uncertainty, which accumulates over long horizons. As illustrated in Fig. \ref{fig:ip_rw}, standard TS rollouts (orange) from a PE model trained on a one-dimensional random walk (blue) significantly overestimate the system's stochasticity. To address this, \cite{frauenknechtrollouts2025} propose Infoprop, a rollout mechanism that uses a maximum likelihood estimation to remove epistemic uncertainty from the predictive distribution and an information-theoretic approach to model rollouts to track data corruption over long horizons. The resulting Infoprop distribution (purple) closely approximates the environment behavior.

 \begin{figure*}[tb]
     \centering
     \begin{minipage}[b]{0.32\textwidth}
         \centering
         \includegraphics[width=\textwidth]{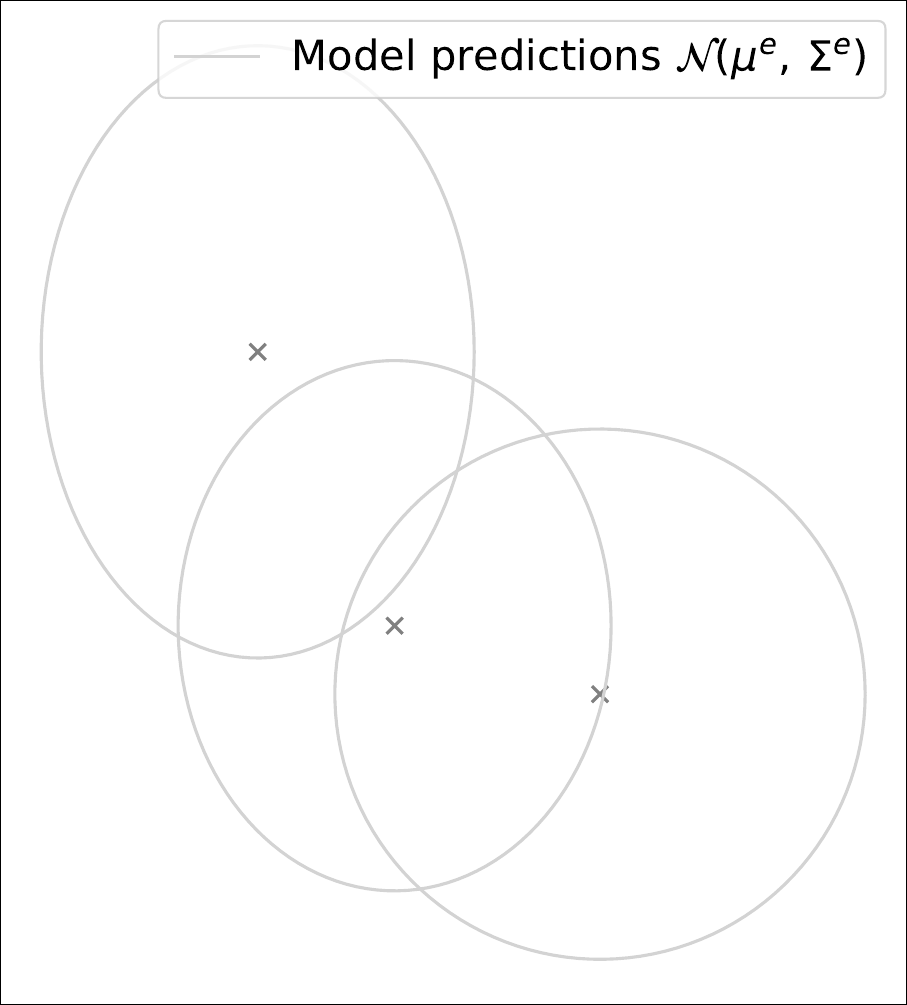}
         \par\smallskip
         (a) Trajectory Sampling Distribution
     \end{minipage}
     \hfill
     \begin{minipage}[b]{0.32\textwidth}
         \centering
         \includegraphics[width=\textwidth]{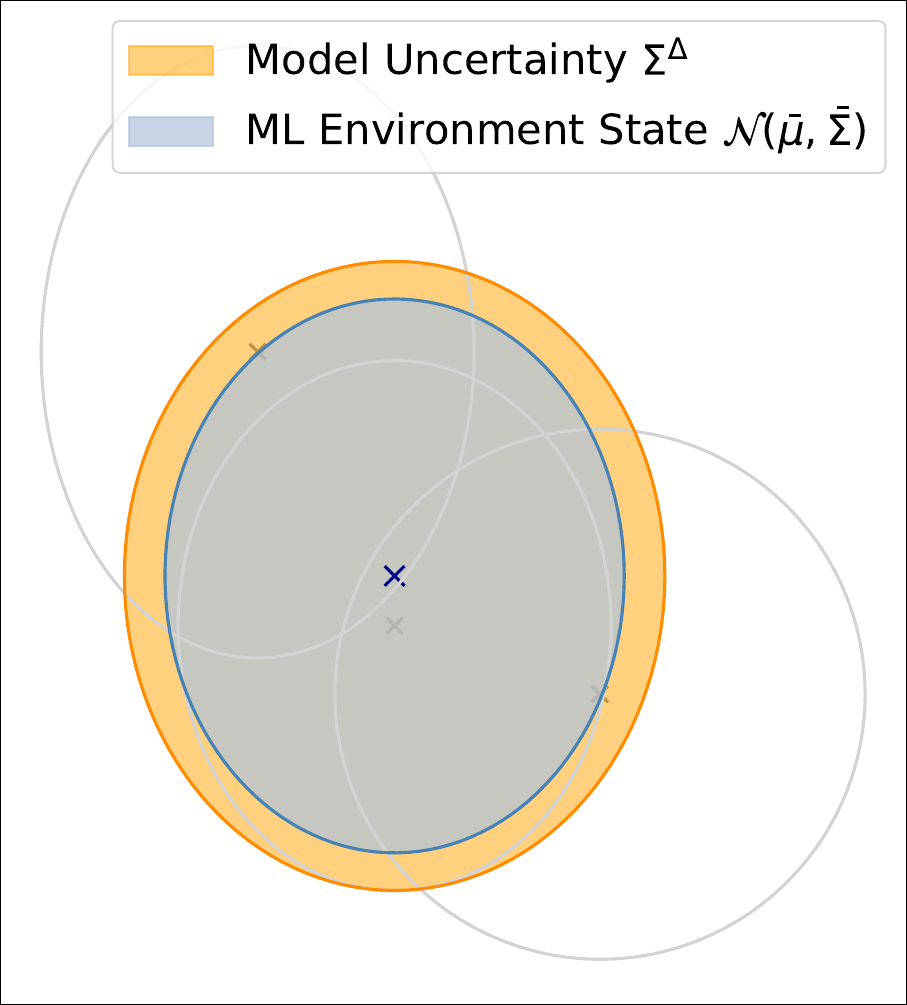}
         \par\smallskip
         (b) Maximum Likelihood Distribution
     \end{minipage}
     \hfill
     \begin{minipage}[b]{0.32\textwidth}
         \centering
         \includegraphics[width=\textwidth]{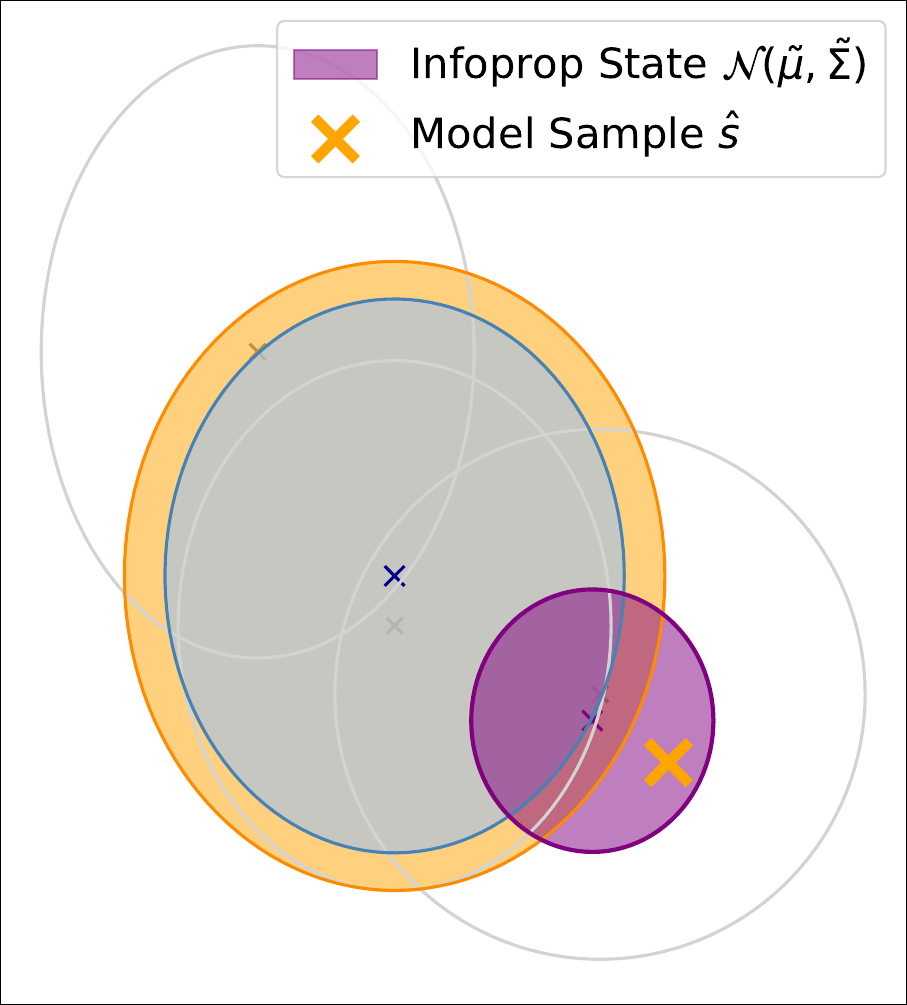}
         \par\smallskip
         (c) Infoprop Distribution
     \end{minipage}

     \caption{Constructing the Infoprop Distribution. \textit{ Based on the PE predictive distribution, used for TS rollouts (a), a ML estimate of the environment distribution and an estimate of epistemic stochasticity are computed (b). Given a random sample from the PE distribution, the corresponding environment state is estimated. This yields the Infoprop distribution that becomes increasingly stochastic in the face of model uncertainty (c). Ellipses indicate one standard deviation.}}
     \label{fig:ip_construction}
     \vspace{\figmargin}
\end{figure*}

The Gaussian mixture predictive distribution resulting from TS rollouts \eqref{eq:ts_dist}, is depicted in Fig. \ref{fig:ip_construction}a. 
 Within the sufficiently certain set $\mathcal{E}$, we assume the PE model to be a consistent and unbiased estimator. Thus, we interpret the individual PNN predictions as noisy observations of a common ground truth, namely the predictive distribution of the environment $p$. Inspired by the field of sensor fusion, we obtain a maximum likelihood (ML) estimate of the environment distribution using covariance intersection fusion \cite{Julier01}. This yields
\begin{equation}
    \Bar{p}(\cdot | s_t, a_t) = \mathcal{N}(\cdot|\bar{\mu}(s_t, a_t), \bar{\Sigma}(s_t, a_t))
    \label{eq:ml_dist}
\end{equation}
with covariance $\Bar{\Sigma}(s_t, a_t) =  (\frac{1}{E} \sum_{e=1}^E \Sigma^e(s_t, a_t)^{-1})^{-1}$ and mean $\bar{\mu}(s_t, a_t) = \Bar{\Sigma}(s_t, a_t)(\frac{1}{E} \sum_{e=1}^E (\Sigma^e(s_t, a_t)^{-1}\mu^e(s_t, a_t))$, depicted in Fig. \ref{fig:ip_construction}b. While sampling from $\bar{p}$ yields a close approximation of the environment dynamics, represented by the purple distribution of  Fig. \ref{fig:ip_rw}, it does not yet provide information about the trustworthiness of the ML estimate.

In other words, depending on the predictive accuracy of the PE model, we lose information about the environment when querying the model. Further, we need to estimate how this problem accumulates along model-based rollouts. Introducing an estimate for epistemic stochasticity
$\Sigma^\Delta(s_t, a_t) = \frac{1}{E} \sum_{e=1}^E (\mu^e(s_t, a_t) - \bar{\mu}(s_t, a_t)) (\mu^e(s_t, a_t) - \bar{\mu}(s_t, a_t))^\top$, shown in Fig. \ref{fig:ip_construction}b, allows us to interpret model rollouts as communication through a noisy channel.

Here, the environment dynamics with aleatoric stochasticity represent the signal we aim to propagate and the epistemic stochasticity represents the channel noise that corrupts the data. Given a sample from the model $\hat{s}_{t+1} \sim \hat{p}(\cdot|s_t, a_t)$ that represents a realization of both the aleatoric and the epistemic stochasticity, we aim to infer the corresponding environment state. This yields the Infoprop state, distributed according to
\begin{equation}
    \tilde{p}(\cdot| s_t, a_t) = \mathcal{N}(\cdot|\tilde{\mu}(s_t, a_t), \tilde{\Sigma}(s_t, a_t))
    \label{eq:ip_dist}
\end{equation}
with $\tilde{\mu}(s_t, a_t) = \bar{\mu}(s_t, a_t) + K(s_t, a_t) (\hat{s}_{t+1} - \bar{\mu}(s_t, a_t))$, $\tilde{\Sigma}(s_t, a_t) = (I-K(s_t, a_t))\bar{\Sigma}(s_t, a_t)$, and $K(s_t, a_t) = \bar{\Sigma}(s_t, a_t) (\bar{\Sigma}(s_t, a_t) + \Sigma^\Delta(s_t, a_t))^{-1}$. Fig. \ref{fig:ip_construction}c illustrates the Infoprop distribution corresponding to a certain model sample.

For an accurate model with $\Sigma^\Delta(s_t, a_t) = 0$, the mapping between model and environment state is clear as reflected by the predictive covariance $\tilde{\Sigma}(s_t, a_t) = 0$, however, the mapping becomes increasingly ambiguous as epistemic stochasticity increases. Consequently, the entropy of $\tilde{p}$ captures epistemic uncertainty, making $u_{\mathrm{epi}}(s_t, a_t) = \mathbb{H}(\tilde{p}(\cdot| s_t, a_t))$ a valid choice.
In particular, the sum of entropies along a rollout corresponds to the information lost about the environment by querying the model. We formulate sufficiently certain paths
\begin{equation}
    \mathcal{P}^t := \Big\{ (s_k, a_k)_{k=0}^t \in (\mathcal{S} \times \mathcal{A})^t \Big|  \sum_{k=0}^t\mathbb{H}(\tilde{p}(\cdot| s_k, a_k)) \leq \kappa_2 \Big\}
    \label{eq:cert_paths}
\end{equation}
and force model usage to stay within these paths to avoid long-horizon data corruption, in addition to keeping the rollouts within the certain area as per \eqref{eq:cert_set}. The accumulation termination in Fig. \ref{fig:fig1} indicates the restriction of model usage to these sufficiently certain paths.
Finally, rollouts under $\tilde{p}$ are equally distributed to those under the ML estimate $\bar{p}$ in expectation, yielding a close approximation of the environment. For a detailed discussion of Infoprop distribution properties see \cite{frauenknechtrollouts2025}.

\section{Leveraging Uncertainty Awareness}
\label{sec:uncertainty_usage}
Integrating the insights of Section \ref{sec:uncertainty_quantification} into state-of-the-art Dyna-style MBRL algorithms \cite{janner_when_2019-1} effectively mitigates model exploitation, yielding substantial improvements in data efficiency and asymptotic performance \cite{frauenknecht2024trust, frauenknechtrollouts2025}. In particular, Infoprop \cite{frauenknechtrollouts2025} allows to plan over substantially increased horizons compared to prior methods. In the following, we present recent applications to hardware learning and safe exploration.

\subsection{Learning on Hardware}
Prominent successes in robot learning \cite{rudin2022learning, kaufmann2023champion} are typically achieved through massively parallelized pretraining of the RL agent in an engineered physics simulator with domain randomization and subsequent deployment on the hardware. While these sim-to-real workflows can yield strong results, they have several downsides. Implementing a sufficiently accurate, yet fast simulator requires substantial engineering effort and becomes increasingly problematic for phenomena that are hard to model from first principles. Further, finding randomization parameters yielding a sufficiently robust agent for a successful transfer to hardware, while not forcing overly conservative behavior, is generally tedious.

Having addressed model exploitation, especially over long prediction horizons, makes MBRL a promising alternative. Restricting model usage to sufficiently certain regions, the learned PE model accurately reflects environment behavior. Further, simulating with a PE model is fast and easily parallelized. Finally, the aleatoric stochasticity provides the agent with dynamics randomization that is reflected in the data obtained from the real system. As more data is collected, certain regions typically grow, yielding a more powerful simulator.

\begin{figure}[tb]
\centerline{\includegraphics[width=\columnwidth]{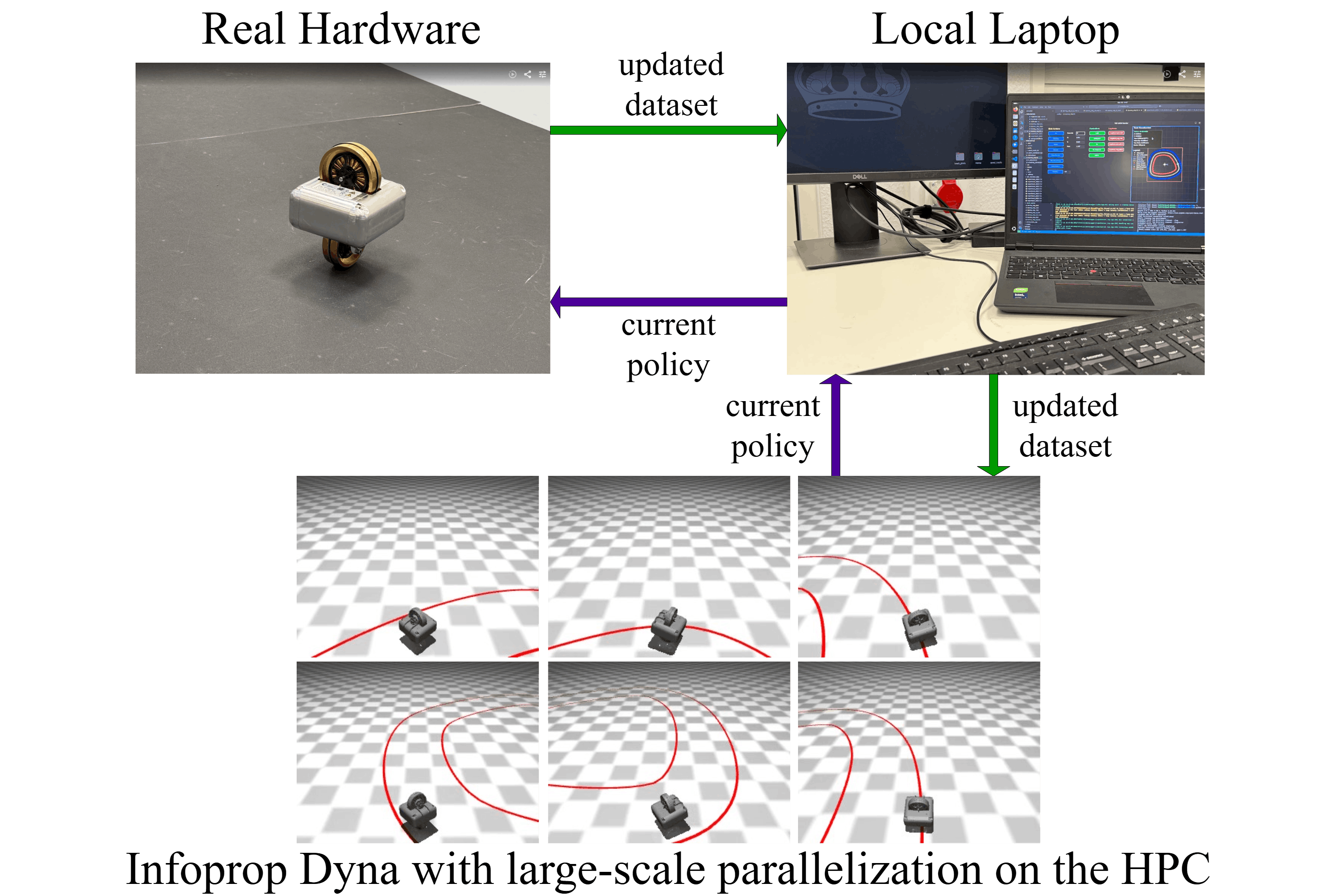}}
\caption{(from \cite{subhasish2026}) Racing the Mini Wheelbot with MBRL: \textit{Data is collected under the current policy directly on hardware. The uncertainty-aware PE model is leveraged as an accurate data-driven simulator that improves with more data.}}
\label{fig:wheelbot_racing}
\vspace{\figmargin}
\end{figure}

Following this idea, \cite{subhasish2026} demonstrate racing on the Mini Wheelbot \cite{hose2025mini} with MBRL using the Infoprop mechanism, as depicted in Fig. \ref{fig:wheelbot_racing}. The Mini Wheelbot is an underactuated unicycle robot with fast and unstable dynamics, requiring a high control frequency since the RL agent directly controls motor torques. Data collected directly on the hardware is communicated to a local laptop where it is added to the overall dataset. Given this data, the PE model is retrained and used for parallelized simulations on a high-performance computing (HPC) cluster. For illustration purposes, simulation states are rendered in MuJoCo \cite{Todorov} while the PE model serves as physics engine. The resulting agent is deployed on the hardware for the next round of data collection to improve the model.

This implementation yields a policy that substantially outperforms a nonlinear model predictive control (MPC) baseline \cite{hose2025approximate} after only 11 minutes of real-world interaction, illustrating the capabilities of uncertainty-aware MBRL.

\subsection{Safe Exploration}
The real-world deployment of RL agents is further hindered by their unsafe exploration behavior that can cause damage to the hardware and endanger operators. Scalable approaches to exploration safety such as constrained MDPs \cite{achiam_constrained_2017, as_actsafe_2025} encourage failures to be below a cost budget in expectation, which is insufficient in many scenarios. On the other hand, safety filters providing rigorous guarantees \cite{wabersich2023data, ames_control_2017, bansal_hamilton-jacobi_2017} typically require a prohibitive amount of domain knowledge or do not scale to many problems of interest.

Safety certification requires evaluating long-term behavior and detecting potential failures. By leveraging Infoprop and adding the failure scenario of leaving  $\mathcal{E}$, we can conduct this evaluation safely within the PE model. Following this intuition, two orthogonal approaches to safe exploration using uncertainty-aware MBRL were recently proposed.

\paragraph{Dyna-style Safety Augmented Reinforcement Learning (Dyna-SAuR) \cite{eisele2026}}
Viability-based safety filters such as control barrier functions \cite{ames_control_2017} can be expressed as an RL policy that separates the action space $\mathcal{A}$ into viable actions and unviable actions, given the current state \cite{Lavanakul24a}. Viable actions keep the system safe, while unviable actions eventually lead to failures.

\begin{figure}[tb]
\centerline{\includegraphics[width=\columnwidth]{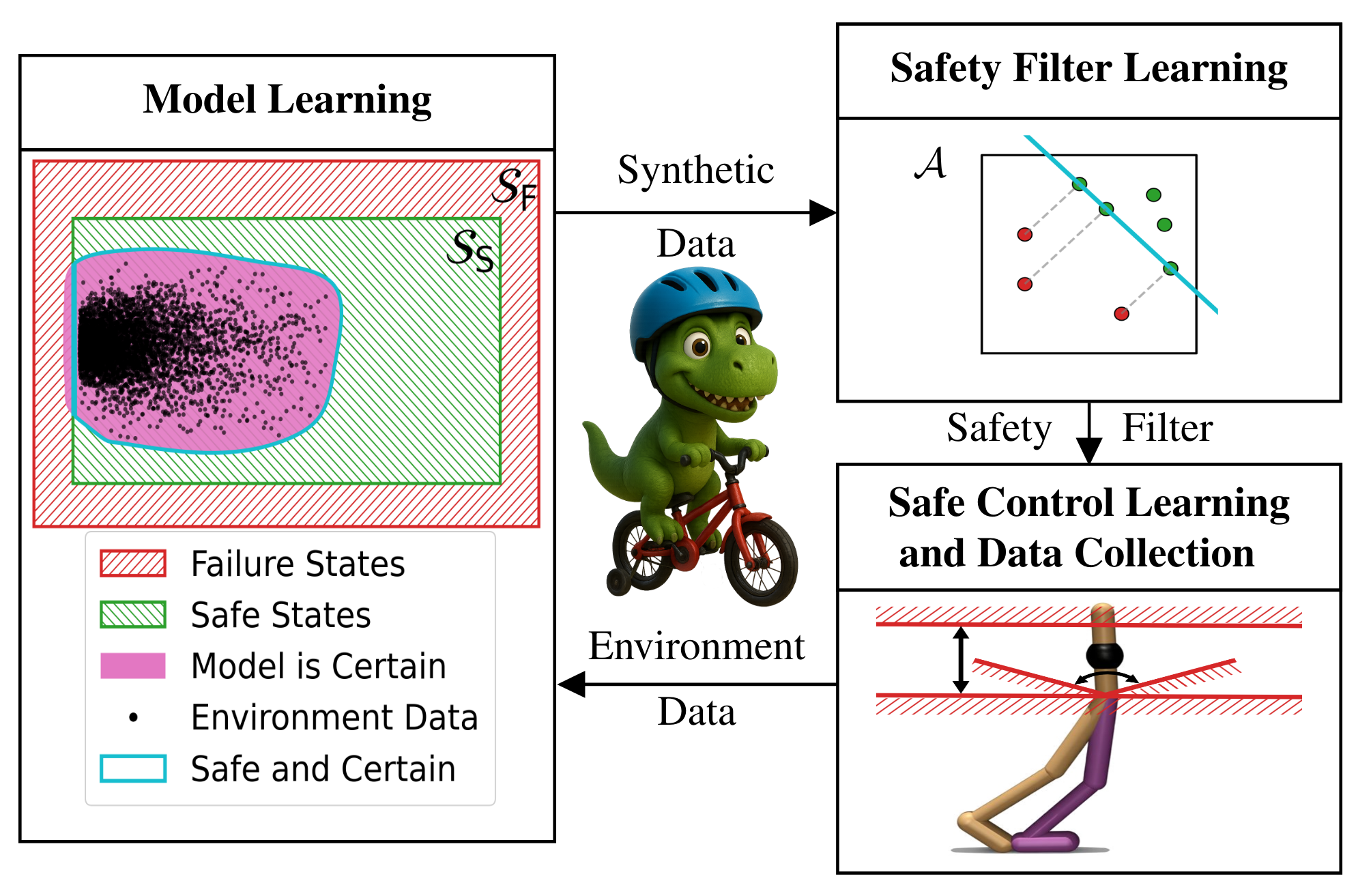}}
\caption{(Adapted from \cite{eisele2026}) Dyna-SAuR \cite{eisele2026}. \textit{An uncertainty-aware dynamics model is used to train a safety filter that avoids both failures and uncertain regions of the model. The filter is used to safely learn a control policy in the environment. The collected data is used to improve the dynamics model, which expands the certain area and reduces conservatism.}}
\label{fig:dynasaur}
\vspace{\figmargin}
\end{figure}

Following this insight, Dyna-SAuR learns both a control and a filter policy via MBRL, as depicted in Fig. \ref{fig:dynasaur}. Given a definition of failure $\mathcal{S}_\mathrm{F}$ and initial data to train the PE model, a filter policy learns to map actions taken by the control policy to the closest viable action, such that failure states in $\mathcal{S}_\mathrm{F}$ as well as uncertain areas are avoided. This requires simulating long-horizon Infoprop trajectories to evaluate whether failure or uncertainty occurs and to classify actions as viable or unviable. Given this filter policy, the control policy can safely interact with the system to collect further data.
Thus, filter policies become less restrictive over time as the certain area grows.

As a sampling-based approximation to viability-based safety filters, Dyna-SAuR does not yield rigorous guarantees. However, it scales to systems that thus far could only be addressed by constrained MDP approaches and reduces failures by two orders of magnitude compared to these methods \cite{eisele2026}.

\paragraph{Uncertainty-Aware Predictive Safety Filter (UPSi) \cite{frauenknecht2026}} An alternative to an explicit safety filter, as in Dyna-SAuR, is leveraging decision time planning similar to MPC \cite{rawlings_model_2024} to avoid failures. These predictive safety filters (PSF) \cite{wabersich_linear_2018} verify whether an MPC scheme can find a constraint-satisfying solution for the current RL action, and adapt it if necessary. 
Despite successful hardware applications \cite{bejarano_safety_2025}, PSFs typically rely on model classes from MPC literature such as engineering models \cite{bejarano_safety_2025} or Gaussian processes \cite{koller_learning-based_2018} that either require substantial domain knowledge or scale poorly with data. In contrast, state-of-the-art MBRL relies on NN architectures that allow to model high dimensional, nonlinear dynamics from large datasets and with minimal prior knowledge. Thus, integrating PE models into PSFs holds the promise to scale PSFs to systems where RL-based control is most relevant. Prior work \cite{gronauer_reinforcement_2024}, however, lacks a careful handling of model uncertainty and therefore loses rigorous guarantees.

Building on the ML estimate of the predictive distribution in Infoprop and by making model certainty a general nonlinear constraint of the MPC problem, UPSi develops a rigorous computation of reachable sets based on a PE model. Integrating a simplified version of UPSi into Dyna-style MBRL \cite{janner_when_2019-1}, yields a substantial reduction of failures during training \cite{frauenknecht2026}.

\section{Concluding Remarks and Outlook}

Introducing uncertainty-aware mechanisms into MBRL can effectively mitigate model exploitation, substantially enhancing its capabilities. Besides improving data efficiency and performance in standard benchmarks \cite{frauenknecht2024trust, frauenknechtrollouts2025}, the capability to perform long and accurate model rollouts opens several promising avenues for MBRL approaches.

While recent success in side-stepping common sim-to-real workflows \cite{subhasish2026}, and scaling safety filter approaches to high dimensional, nonlinear systems \cite{eisele2026, frauenknecht2026} mark an exciting beginning, we consider the applicability of uncertainty-aware MBRL methods much broader. 
The presented methods yield access to a data driven simulator that can clearly communicate its capability to approximate the environment over long horizons and that improves over time as more data is collected. This could benefit fields like model-based planning \cite{Williams2017May, Chua2018}, active exploration \cite{buisson2020actively, sukhijamaxinforl}, and transfer learning \cite{luo2024progressive}, addressing long-standing problems in robot learning.

Further, visual observations and the corresponding latent dynamics model architectures \cite{Hafner2020Apr, hansentd} are a highly relevant area of MBRL. Uncertainty quantification for latent space models is currently underexplored \cite{beckeruncertainty} and their behavior fundamentally differs from models for proprioceptive state representations \cite{berger2026}. Investigating the underlying differences to adapt the presented methods for latent dynamics models is another exciting direction for future research.

\section*{Acknowledgment} We thank Julia Berger, Bastian Leibe, Lukas Kesper, and Henrik Hose for their contributions to this line of research, as well as Christian Fiedler and Pierre-Fran\c{c}ois Massiani  for the fruitful discussions on Infoprop.

\bibliographystyle{IEEEtran}
\bibliography{icra_mbrl}

@InProceedings{janner_when_2019-1,
  author    = {Janner, Michael and Fu, Justin and Zhang, Marvin and Levine, Sergey},
  booktitle = {Adv. in {Neural} {Inf.} {Proc.} {Sys.}},
  title     = {When to {Trust} {Your} {Model}: {Model}-{Based} {Policy} {Optimization}},
  year      = {2019},
}

@InProceedings{frauenknecht_data-efficient_2023,
  author    = {Frauenknecht, Bernd and Ehlgen, Tobias and Trimpe, Sebastian},
  booktitle = {Int. Conf. on {Intelligent} {Transportation} {Systems}},
  title     = {Data-efficient {Deep} {Reinforcement} {Learning} for {Vehicle} {Trajectory} {Control}},
  year      = {2023},
  doi       = {10.1109/ITSC57777.2023.10422451},
  murldate  = {2024-10-15},
}

@Article{frauenknecht2026,
  author  = {Frauenknecht, Bernd and Kesper, Lukas and Mayfrank, Daniel and Hose, Henrik and Trimpe, Sebastian},
  journal = {Reinforcement Learning Conf.},
  title   = {Uncertainty-Aware Predictive Safety Filters for Probabilistic Neural Network Dynamics},
  year    = {2026},
}

@Article{eisele2026,
  author  = {Eisele, Artur and Frauenknecht, Bernd and Solowjow, Friedrich and Trimpe, Sebastian},
  journal = {arxiv preprint},
  title   = {Dyna-Style Safety Augmented Reinforcement Learning: Staying Safe in the Face of Uncertainty},
  year    = {2026},
}

@Article{berger2026,
  author  = {Berger, Julia and Frauenknecht, Bernd and Trimpe, Sebastian and Leibe, Bastian},
  journal = {Reinforcement Learning Conf.},
  title   = {Biased Dreams: Limitations to Epistemic Uncertainty Quatification in Latent Space Models},
  year    = {2026},
}

@Article{subhasish2026,
  author  = {Subhasish, Devdutt and Hose, Henrik and Trimpe, Sebastian},
  journal = {arxiv},
  title   = {Learning to Race in Minutes: Infoprop Dyna on the Mini Wheelbot},
  year    = {2026},
}

@InProceedings{frauenknecht2024trust,
  author    = {Frauenknecht, Bernd and Eisele, Artur and Subhasish, Devdutt and Solowjow, Friedrich and Trimpe, Sebastian},
  booktitle = {Int. Conf. on Mach. Learn.},
  title     = {Trust the model where it trusts itself: model-based actor-critic with uncertainty-aware rollout adaption},
  year      = {2024},
}

@InProceedings{frauenknechtrollouts2025,
  author    = {Frauenknecht, Bernd and Subhasish, Devdutt and Solowjow, Friedrich and Trimpe, Sebastian},
  booktitle = {Int. Conf. on Learn. Rep.},
  title     = {On Rollouts in Model-Based Reinforcement Learning},
  year      = {2025},
}

@Article{beckeruncertainty,
  author  = {Becker, Philipp and Neumann, Gerhard},
  journal = {Transactions on Machine Learning Research},
  title   = {On Uncertainty in Deep State Space Models for Model-Based Reinforcement Learning},
  year    = {2022},
}

@InProceedings{buisson2020actively,
  author    = {Buisson-Fenet, Mona and Solowjow, Friedrich and Trimpe, Sebastian},
  booktitle = {Learning for dynamics and control},
  title     = {Actively learning gaussian process dynamics},
  year      = {2020},
}

@InProceedings{sukhijamaxinforl,
  author    = {Sukhija, Bhavya and Coros, Stelian and Krause, Andreas and Abbeel, Pieter and Sferrazza, Carmelo},
  booktitle = {Int. Conf. on Learning Representations},
  title     = {MaxInfoRL: Boosting exploration in reinforcement learning through information gain maximization},
  year      = {2025},
}

@Article{Tang2025May,
  author    = {Tang, Chen and Abbatematteo, Ben and Hu, Jiaheng and Chandra, Rohan and Mart{\ifmmode\acute{\imath}\else\'{\i}\fi}n-Mart{\ifmmode\acute{\imath}\else\'{\i}\fi}n, Roberto and Stone, Peter},
  journal   = {Annu. Rev. Control Rob. Auton. Syst.},
  title     = {{Deep Reinforcement Learning for Robotics: A Survey of Real-World Successes}},
  year      = {2025},
  publisher = {Annual Reviews},
}

@InProceedings{luo2019,
  author    = {Luo, Yuping and Xu, Huazhe and Li, Yuanzhi and Tian, Yuandong and Darrell, Trevor and Ma, Tengyu},
  booktitle = {Int. Conf. on Learning Representations},
  title     = {Algorithmic Framework for Model-based Deep Reinforcement Learning with Theoretical Guarantees},
  year      = {2019},
}

@InBook{Julier01,
  author  = {Julier, Simon and Uhlmann, Jeffrey},
  title   = {General Decentralized Data Fusion with Covariance Intersection (CI)},
  year    = {2001},
  month   = {06},
  doi     = {10.1201/9781420038545.ch12},
  journal = {Handbook of Multisensor Data Fusion, Theroy and Practice},
}

@Article{Chua2018,
  author  = {Chua, Kurtland and Calandra, Roberto and McAllister, Rowan and Levine, Sergey},
  journal = {Adv. in Neural Information Processing Systems},
  title   = {{Deep Reinforcement Learning in a Handful of Trials using Probabilistic Dynamics Models}},
  year    = {2018},
}

@Article{Feinberg2018Feb,
  author  = {Feinberg, Vladimir and Wan, Alvin and Stoica, Ion and Jordan, Michael I. and Gonzalez, Joseph E. and Levine, Sergey},
  journal = {{Int. Conf. on Machine Learning}},
  title   = {{Model-Based Value Estimation for Efficient Model-Free Reinforcement Learning}},
  year    = {2018},
}

@InCollection{Buckman2018Dec,
  author    = {Buckman, Jacob and Hafner, Danijar and Tucker, George and Brevdo, Eugene and Lee, Honglak},
  booktitle = {{Int. Conf. on Neural Information Processing Systems}},
  title     = {{Sample-efficient reinforcement learning with stochastic ensemble value expansion}},
  year      = {2018},
}

@InCollection{Hafner2019May,
  author    = {Hafner, Danijar and Lillicrap, Timothy and Fischer, Ian and Villegas, Ruben and Ha, David and Lee, Honglak and Davidson, James},
  booktitle = {{Int. Conf. on Machine Learning}},
  publisher = {PMLR},
  title     = {{Learning Latent Dynamics for Planning from Pixels}},
  year      = {2019},
  month     = may,
  journal   = {PMLR},
}

@InProceedings{Hafner2020Apr,
  author    = {Hafner, Danijar and Lillicrap, Timothy and Ba, Jimmy and Norouzi, Mohammad},
  booktitle = {Int. Conf. on Learning Representations},
  title     = {{Dream to Control: Learning Behaviors by Latent Imagination}},
  year      = {2020},
  month     = apr,
}

@InCollection{Nagabandi2018,
  author    = {Nagabandi, Anusha and Kahn, Gregory and Fearing, Ronald S. and Levine, Sergey},
  booktitle = {{Int. Conf. on Robotics and Automation}},
  title     = {{Neural Network Dynamics for Model-Based Deep Reinforcement Learning with Model-Free Fine-Tuning}},
  year      = {2018},
}

@Article{Sutton1991Jul,
  author  = {Sutton, Richard S.},
  journal = {SIGART Bull.},
  title   = {{Dyna, an integrated architecture for learning, planning, and reacting}},
  year    = {1991},
}

@InCollection{Williams2017May,
  author    = {Williams, Grady and Wagener, Nolan and Goldfain, Brian and Drews, Paul and Rehg, James M. and Boots, Byron and Theodorou, Evangelos A.},
  booktitle = {{Int. Conf. on Robotics and Automation}},
  title     = {{Information theoretic MPC for model-based reinforcement learning}},
  year      = {2017},
}

@InCollection{Lakshminarayanan2017Dec,
  author    = {Lakshminarayanan, Balaji and Pritzel, Alexander and Blundell, Charles},
  booktitle = {{Int. Conf. on Neural Information Processing Systems}},
  title     = {{Simple and scalable predictive uncertainty estimation using deep ensembles}},
  year      = {2017},
}

@InCollection{Deisenroth2011Jun,
  author    = {Deisenroth, Marc Peter and Rasmussen, Carl Edward},
  booktitle = {{Int. Conf. on Mach. Learn.}},
  title     = {{PILCO: a model-based and data-efficient approach to policy search}},
  year      = {2011},
}

@InProceedings{gal2016improving,
  author    = {Gal, Yarin and McAllister, Rowan and Rasmussen, Carl Edward},
  booktitle = {Int. Conf. on Mach. Learn.},
  title     = {Improving {PILCO} with {B}ayesian neural network dynamics models},
  year      = {2016},
}

@InCollection{Gu2016Jun,
  author    = {Gu, Shixiang and Lillicrap, Timothy and Sutskever, Ilya and Levine, Sergey},
  booktitle = {{Int. Conf. on Mach. Learn.}},
  title     = {{Continuous Deep Q-Learning with Model-based Acceleration}},
  year      = {2016},
}

@Article{Depeweg2016May,
  author  = {Depeweg, Stefan and Hern{\ifmmode\acute{a}\else\'{a}\fi}ndez-Lobato, Jos{\ifmmode\acute{e}\else\'{e}\fi} Miguel and Doshi-Velez, Finale and Udluft, Steffen},
  journal = {Int. Conf. on Learn. Rep.},
  title   = {{Learning and Policy Search in Stochastic Dynamical Systems with Bayesian Neural Networks}},
  year    = {2017},
}

@InProceedings{hose2025mini,
  author    = {Hose, Henrik and Weisgerber, Jan and Trimpe, Sebastian},
  booktitle = {IEEE Int. Conf. on Robotics and Automation},
  title     = {The Mini Wheelbot: A testbed for learning-based balancing, flips, and articulated driving},
  year      = {2025},
}

@InProceedings{rudin2022learning,
  author    = {Rudin, Nikita and Hoeller, David and Reist, Philipp and Hutter, Marco},
  booktitle = {Conf. on robot learning},
  title     = {Learning to walk in minutes using massively parallel deep reinforcement learning},
  year      = {2022},
}

@Article{kaufmann2023champion,
  author    = {Kaufmann, Elia and Bauersfeld, Leonard and Loquercio, Antonio and M{\"u}ller, Matthias and Koltun, Vladlen and Scaramuzza, Davide},
  journal   = {Nature},
  title     = {Champion-level drone racing using deep reinforcement learning},
  year      = {2023},
  publisher = {Nature Publishing Group UK London},
}

@InProceedings{Lavanakul24a,
  author    = {Lavanakul, Will and Choi, Jason and Sreenath, Koushil and Tomlin, Claire},
  booktitle = {Learning for Dynamics and Control Conference},
  title     = {Safety filters for black-box dynamical systems by learning discriminating hyperplanes},
  year      = {2024},
}

@Article{wabersich2023data,
  author  = {Wabersich, Kim P and Taylor, Andrew J and Choi, Jason J and Sreenath, Koushil and Tomlin, Claire J and Ames, Aaron D and Zeilinger, Melanie N},
  journal = {IEEE Control Systems Magazine},
  title   = {Data-driven safety filters: Hamilton-jacobi reachability, control barrier functions, and predictive methods for uncertain systems},
  year    = {2023},
}

@Article{ames_control_2017,
  author  = {Ames, Aaron D. and Xu, Xiangru and Grizzle, Jessy W. and Tabuada, Paulo},
  journal = {IEEE Transactions on Automatic Control},
  title   = {Control {Barrier} {Function} {Based} {Quadratic} {Programs} for {Safety} {Critical} {Systems}},
  year    = {2017},
}

@InProceedings{koller_learning-based_2018,
  author    = {Koller, Torsten and Berkenkamp, Felix and Turchetta, Matteo and Krause, Andreas},
  booktitle = {{Conf.} on {Decision} and {Control}},
  title     = {Learning-{Based} {Model} {Predictive} {Control} for {Safe} {Exploration}},
  year      = {2018},
  month     = {.},
  note      = {.},
  murldate  = {.},
}

@InProceedings{gronauer_reinforcement_2024,
  author    = {Gronauer, Sven and Haider, Tom and Schmoeller da Roza, Felippe and Diepold, Klaus},
  booktitle = {{Int.} {Conf.} on {Auton.} {Agents} and {Multiagent} {Syst.}},
  title     = {Reinforcement {Learning} with {Ensemble} {Model} {Predictive} {Safety} {Certification}},
  year      = {2024},
}

@article{hose2025approximate,
  title={Approximate nonlinear model predictive control with safety-augmented neural networks},
  author={Hose, Henrik and K{\"o}hler, Johannes and Zeilinger, Melanie N and Trimpe, Sebastian},
  journal={IEEE Transactions on Control Systems Technology},
  year={2025},
  publisher={IEEE}
}

@InCollection{Todorov,
  author    = {Todorov, Emanuel and Erez, Tom and Tassa, Yuval},
  booktitle = {{Int. Conf. on Int. Rob. and Sys.}},
  title     = {{MuJoCo: A physics engine for model-based control}},
  year      = {2012},
}

@Misc{as_actsafe_2025,
  author     = {As, Yarden and Sukhija, Bhavya and Treven, Lenart and Sferrazza, Carmelo and Coros, Stelian and Krause, Andreas},
  title      = {{ActSafe}: {Active} {Exploration} with {Safety} {Constraints} for {Reinforcement} {Learning}},
  year       = {2025},
  abstract   = {Reinforcement learning (RL) is ubiquitous in the development of modern AI systems. However, state-of-the-art RL agents require extensive, and potentially unsafe, interactions with their environments to learn effectively. These limitations confine RL agents to simulated environments, hindering their ability to learn directly in real-world settings. In this work, we present ActSafe, a novel model-based RL algorithm for safe and efficient exploration. ActSafe learns a well-calibrated probabilistic model of the system and plans optimistically w.r.t. the epistemic uncertainty about the unknown dynamics, while enforcing pessimism w.r.t. the safety constraints. Under regularity assumptions on the constraints and dynamics, we show that ActSafe guarantees safety during learning while also obtaining a near-optimal policy in finite time. In addition, we propose a practical variant of ActSafe that builds on latest model-based RL advancements and enables safe exploration even in high-dimensional settings such as visual control. We empirically show that ActSafe obtains state-of-the-art performance in difficult exploration tasks on standard safe deep RL benchmarks while ensuring safety during learning.},
  doi        = {10.48550/arXiv.2410.09486},
  keywords   = {Computer Science - Machine Learning, Computer Science - Robotics},
  murldate   = {2025-04-23},
  publisher  = {arXiv},
  shorttitle = {{ActSafe}},
  _url        = {http://arxiv.org/abs/2410.09486},
}

@InProceedings{achiam_constrained_2017,
  author    = {Achiam, Joshua and Held, David and Tamar, Aviv and Abbeel, Pieter},
  booktitle = {{Int.} {Conf.} on {Machine} {Learning}},
  title     = {Constrained {Policy} {Optimization}},
  year      = {2017},
}

@InProceedings{bansal_hamilton-jacobi_2017,
  author    = {Bansal, Somil and Chen, Mo and Herbert, Sylvia and Tomlin, Claire J.},
  booktitle = {Conf. on {Decision} and {Control}},
  title     = {Hamilton-{Jacobi} reachability: {A} brief overview and recent advances},
  year      = {2017},
}

@Book{rawlings_model_2024,
  author = {Rawlings, James B. and Diehl, Moritz M. and Mayne, David Q.},
  title  = {Model {Predictive} {Control}: {Theory}, {Computation}, and {Design}},
  year   = {2024},
}

@InProceedings{wabersich_linear_2018,
  author    = {Wabersich, Kim P. and Zeilinger, Melanie N.},
  booktitle = {{Conf.} on {Decision} and {Control}},
  title     = {Linear {Model} {Predictive} {Safety} {Certification} for {Learning}-{Based} {Control}},
  year      = {2018},
}

@Article{bejarano_safety_2025,
  author  = {Bejarano, Federico Pizarro and Brunke, Lukas and Schoellig, Angela P.},
  journal = {Robotics and Automation Letters},
  title   = {Safety {Filtering} {While} {Training}: {Improving} the {Performance} and {Sample} {Efficiency} of {Reinforcement} {Learning} {Agents}},
  year    = {2025},
}

@InProceedings{hansentd,
  author    = {Hansen, Nicklas and Su, Hao and Wang, Xiaolong},
  booktitle = {Int. Conf. on Learn. Rep.},
  title     = {TD-MPC2: Scalable, Robust World Models for Continuous Control},
  year      = {2024},
}

@article{Li2025Apr,
	author = {Li, Chenhao and Krause, Andreas and Hutter, Marco},
	title = {{Uncertainty-Aware Robotic World Model Makes Offline Model-Based Reinforcement Learning Work on Real Robots}},
	journal = {arXiv},
	year = {2025},
	month = apr,
	eprint = {2504.16680},
	doi = {10.48550/arXiv.2504.16680}
}

@InProceedings{Yu2020,
  author    = {Yu, Tianhe and Thomas, Garrett and Yu, Lantao and Ermon, Stefano and Zou, James Y and Levine, Sergey and Finn, Chelsea and Ma, Tengyu},
  booktitle = {Adv. in Neural Information Processing Systems},
  title     = {MOPO: Model-based Offline Policy Optimization},
  year      = {2020},
}

@Article{luo2024progressive,
  author    = {Luo, Yongkang and Li, Wanyi and Wang, Peng and Duan, Haonan and Wei, Wei and Sun, Jia},
  journal   = {IEEE Trans. on Cognitive and Developmental Systems},
  title     = {Progressive transfer learning for dexterous in-hand manipulation with multifingered anthropomorphic hand},
  year      = {2024},
  number    = {.},
  pages     = {.},
  volume    = {.},
  publisher = {IEEE},
}
\end{document}